\documentclass[10pt,twocolumn,letterpaper]{article}

\usepackage[pagenumbers]{twocol}

\usepackage[dvipsnames]{xcolor}

\definecolor{myblue}{rgb}{0.21,0.49,0.74}
\usepackage[pagebackref,breaklinks,colorlinks,citecolor=myblue]{hyperref}

\usepackage{xspace}
\makeatletter
\DeclareRobustCommand\onedot{\futurelet\@let@token\@onedot}
\def\@onedot{\ifx\@let@token.\else.\null\fi\xspace}

\def\eg{{e.g}\onedot} 
\def\ie{{i.e}\onedot}

\makeatother

\usepackage{bbm}
\usepackage{xfp}
\usepackage[np,autolanguage]{numprint}
\nprounddigits{1}

\usepackage{algorithm,algorithmic}
\usepackage{stmaryrd}

\newcommand{\method}{ELSA\xspace}
\newcommand{\methodfullname}{Efficient Layer Sparsification Approach\xspace}

\title{\method{}: Partial Weight Freezing for Overhead-Free Sparse Network Deployment}

\author{Paniz Halvachi \\
Sharif University of Technology\!\!\!\!
 \and Alexandra Peste \qquad Dan Alistarh\qquad  Christoph H{.} Lampert \\
Institute of Science and Technology Austria (ISTA)}

\newcommand{\mean}[3]{\fpeval{((#1)+(#2)+(#3))/3}}
\newcommand{\std}[3]{\fpeval{sqrt((\mean{(#1)^2}{(#2)^2}{(#3)^2})-(\mean{#1}{#2}{#3})^2)}}
\newcommand{\meann}[4]{\fpeval{((#1)+(#2)+(#3)+(#4))/4}}
\newcommand{\stdd}[4]{\fpeval{sqrt((\meann{(#1)^2}{(#2)^2}{(#3)^2}{(#4)^2})-(\meann{#1}{#2}{#3}{#4})^2)}}

\newcommand{\meanstd}[3]{\np{\mean{#1}{#2}{#3}}\pm\np{\std{#1}{#2}{#3}}}
\newcommand{\meannstdd}[4]{\np{\meann{#1}{#2}{#3}{#4}}\pm\np{\stdd{#1}{#2}{#3}{#4}}}

\newcommand{\myparagraph}[1]{\medskip\noindent\textbf{#1}\quad}

\begin{document}
\maketitle
\begin{abstract}
    We present \method, a practical solution for creating deep networks that can easily be deployed at different levels of sparsity. 
    The core idea is to embed one or more sparse networks within a single dense network as a proper subset of the weights.
    At prediction time, any sparse model can be extracted  effortlessly simply be zeroing out weights according to a predefined mask. 
    \method is simple, powerful and highly flexible. 
    It can use essentially any existing technique for network sparsification and network training. 
    In particular, it does not restrict the loss function, 
    architecture or the optimization technique. 
    Our experiments show that \method's advantages of flexible deployment comes with no or just a negligible reduction in prediction quality compared to the standard way of using multiple sparse networks that are trained and stored independently. 
\end{abstract}
    
\section{Introduction}\label{sec:intro}
Deep learning has revolutionized a myriad of domains, but the high 
computational and memory cost of large models remains a hurdle. 
Network sparsification techniques have been explored to mitigate 
these issues. However, most existing approaches focus only on creating 
individual sparse models with as-good-as-possible accuracy-vs-sparsity 
trade-off, but they do not consider the ease of their practical deployment. 

As an illustrative example, consider the situation of a 
smartphone app that runs a deep network for object detection. 
Depending on the phone hardware, networks of different 
sizes should to be used to ensure real-time capabilities. 
Traditionally, there are two ways to achieve this:
either, the app binary already contains all possible 
networks and it chooses an appropriate one at execution 
time. 
This is efficient and easy in deployment, but it has the 
disadvantage of an excessively large initial download. 
The app size will grow linearly with the number of possible 
networks, even though only a single one will actually be 
used in the end. 
Alternatively, the app binary would not contain the weights
for any network. Instead, it decides on a suitable network 
size when it is started for the first time, and downloads 
the weights for this network from an online resource. 
This, however, leads to a worse user experience. The first 
start of the app will incur a long latency and unexpected 
large data transmission. It might also trigger security 
concerns if an app downloads large files from an external 
server. 
\begin{figure}[t]
\centerline{\includegraphics[width=.6\columnwidth]{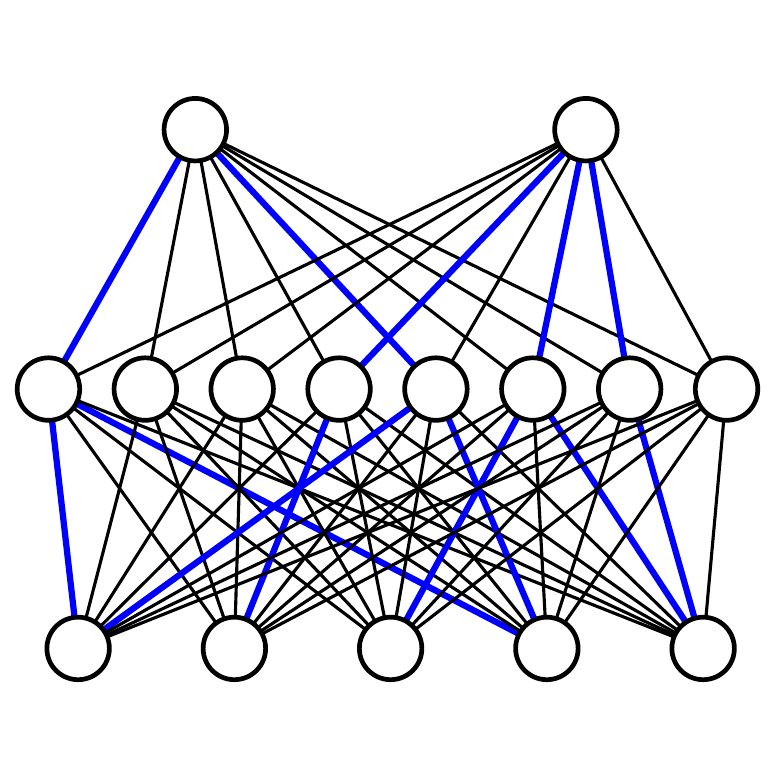}}
\caption{Illustration of an \method-Net: the weights of a trained sparse 
network (blue) are embedded as a proper subset within the weights of a 
dense network. At prediction time, the complete dense network can be 
run as is, or the sparse network can be extracted effortlessly by setting 
all other (non-blue) weights to $0$. The extracted sparse network is 
identical to the one originally embedded, allowing for state-of-the-art 
accuracy without the need for fine-tuning or other adjustments.}\label{fig:teaser}
\end{figure}
More effective than the above two cases, would be a
setup in which the app binary contains just one network, 
from which smaller and more efficient models could be 
constructed on-the-fly to match the device hardware. 
In the context of network sparsification, this 
task is known as \emph{one-shot pruning}.

In this work, \textbf{we introduce \emph{\method (\methodfullname)}, 
a technique that allows one-shot pruning of deep 
networks at deployment time to a degree not achieved 
before.}
\method is, in fact, not a new network sparsification technique, 
but a new way of constructing, storing, and retrieving  
sparse networks, where the individual networks are produced 
using any existing technique for network sparsification.
The core idea is illustrated in Figure~\ref{fig:teaser}: 
the weights of a trained sparse network (blue) are embedded 
as a proper subset within the weights of a dense network. 
At prediction time, the complete dense network can be run as 
is, or the sparse network can be extracted effortlessly by 
setting all other (non-blue) weights to $0$. The extracted sparse 
network is identical to the one originally embedded, allowing for 
state-of-the-art accuracy without the need for fine-tuning 
or other adjustments. 

\begin{figure*}[t]
    \centering
    \subfloat[intial dense network]{\fbox{\includegraphics[width=.18\textwidth]{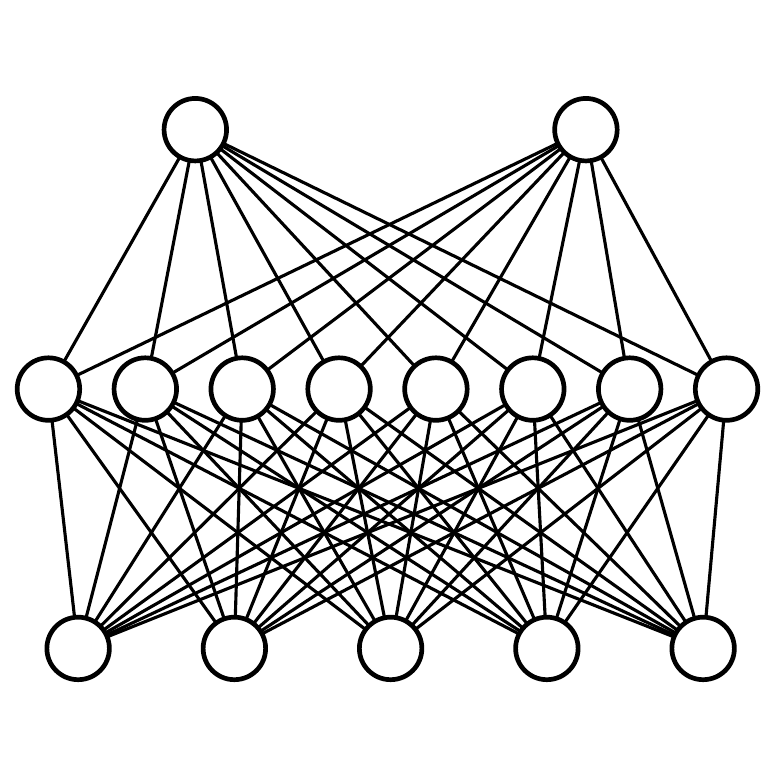}}}\ 
    \subfloat[sparsified network]{\fbox{\includegraphics[width=.18\textwidth]{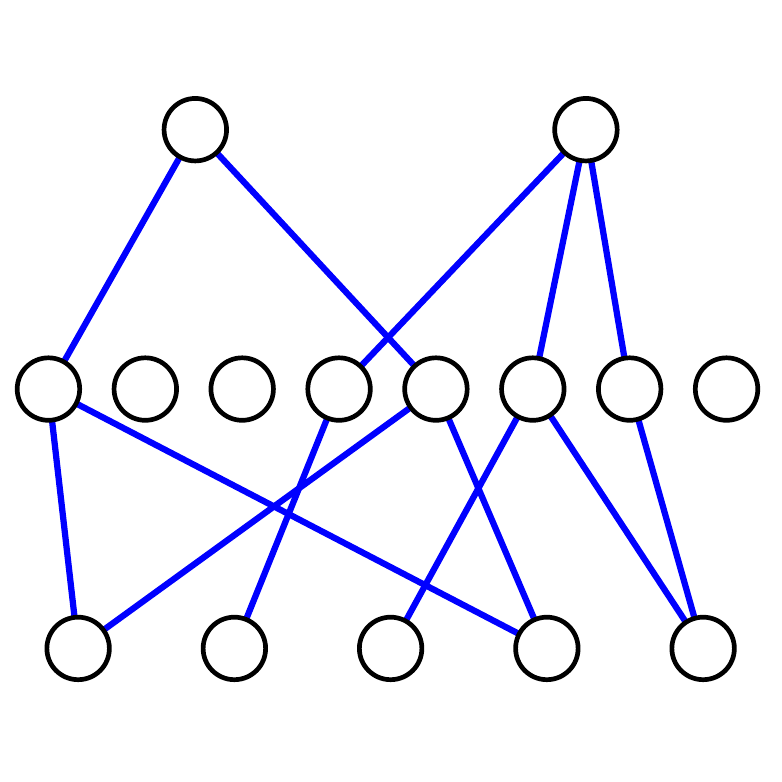}}}\ 
    \subfloat[dense network with  embedded sparse network]{\fbox{\includegraphics[width=.18\textwidth]{images/matryoshka_network.pdf}}}\ 
    \subfloat[sparse network with embedded first sparse network]{\fbox{\includegraphics[width=.18\textwidth]{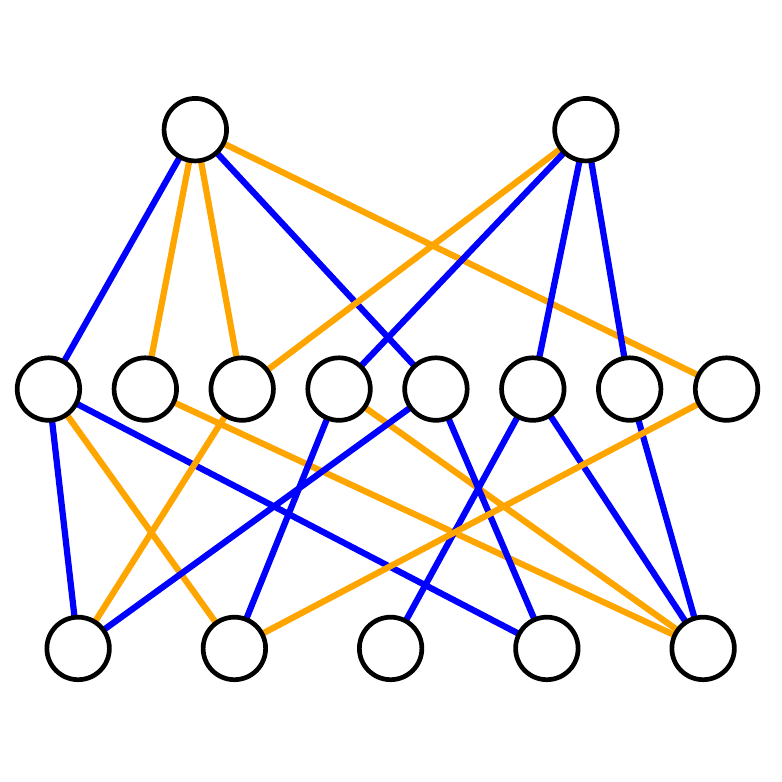}}}\ 
    \subfloat[dense network with two embedded sparse networks]{\fbox{\includegraphics[width=.18\textwidth]{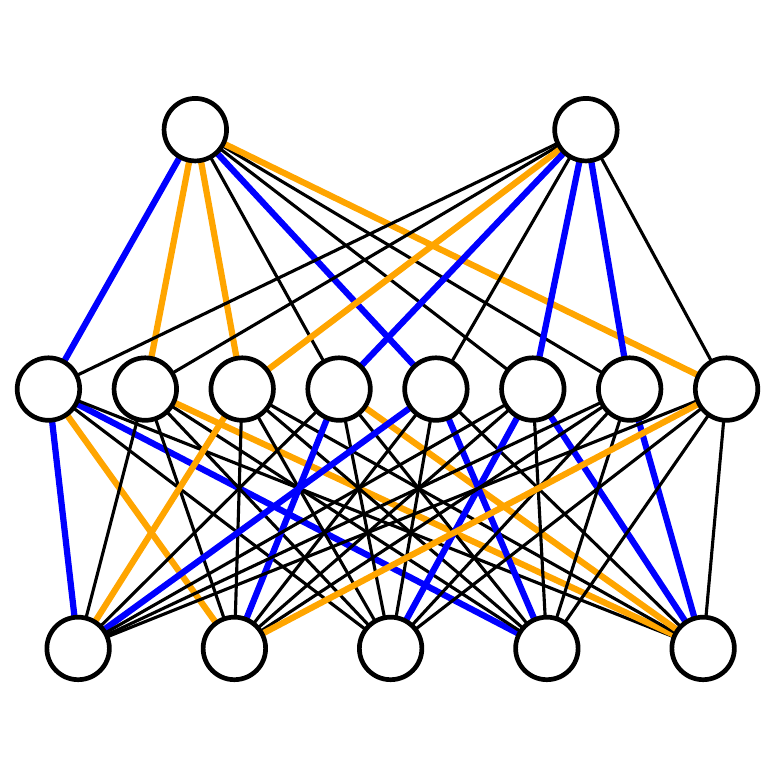}}}
    \caption{Illustration of \emph{\method} (a)--(c) and \emph{multi-level \method} (a)--(e). }\label{fig:multilevel}
\end{figure*}
By iterating the above procedure, it is also possible to embed 
multiple models, of different sparsity levels, within a single 
dense one, thereby providing a choice between different sparsity 
levels at prediction time. 

\textbf{Networks produced by \method (\method-Nets) can have 
arbitrary architectures, and they can be trained, evaluated, 
and deployed using standard techniques with at most minor 
adjustments.
At prediction time, they can be used either in dense or in 
sparse form with negligible overhead. }

\section{Method}\label{sec:method}
In this section we first introduce the \method-method 
of storing a sparse network within a dense one. 
We then discuss a multi-level extension that allows 
creating and storing multiple networks of different 
sparsity levels still within a single dense network.

\subsection{\method} 
At the core of \method lies the observation that the weights 
of a sparse networks can be embedded inside a dense one, 
and later recovered without loss of quality. %
While previously proposed to prevent catastrophic forgetting in 
continual multi-task learning~\cite{mallya2018packnet,srivastava2019adaptive}, 
we instantiate the idea in the context of one-shot network pruning.

Its working principle is illustrated in Figure~\ref{fig:multilevel} (a)--(c).
The starting point is an ordinary dense network~(a), typically pretrained to high accuracy. 
To this, one applies an arbitrary \emph{sparsification} method, resulting in a sparse 
network~(b).
Now, the subset of weights that belong to the sparse network are \emph{frozen} 
(illustrated in blue). This means that their values will be prevented from changing 
in subsequent steps. 
Then, a \emph{densification} step is performed. This trains the non-frozen weights
(which previously all had value $0$), thereby again ending up with a dense network (c). 

A brief analysis shows the rationality of the process: because the choice 
of sparsification method is  discretionary, the sparse network 
can be of state-of-the-art quality. 
The dense network contains the sparse one as a fixed subset, but it has additional 
learning capacity. As such, one can expect the resulting dense network to be of 
higher accuracy than the sparse one, ideally recovering or even exceeding the 
accuracy of the initial dense network. 

The sparse network can be extracted effortlessly from the dense one, simply by 
setting a predefined set of network weights to zero. 
The resulting sparse network is identical to the originally trained one, so its
efficiency and quality are known a priori, and no further finetuning or other 
adjustments are needed to give it high accuracy. 
The latter aspect is the main difference to prior one-shot pruning works: these
are also able to create sparse networks, but cannot guarantee their quality. 
Typically, they also require some additional processing and data to achieve 
satisfying accuracy (see our discussion in Section~\ref{sec:related}).

\begin{algorithm}[t]
\caption{\texttt{\method}}\label{alg:single}
\begin{algorithmic}[1]
\INPUT network with weights $\theta\in\mathbb{R}^D$
\INPUT target sparsity level $\alpha$
\INPUT\textbf{(optional)} frozen/learnable mask $M\in\{0,1\}^D$ % % frozen\,(0) vs learnable\,(1)
\SHORTIF{no $M$ provided}{$M\leftarrow \mathbf{1}\in\mathbb{R}^D$}
\STATE $\theta_{\text{sparse}} \leftarrow \texttt{sparsify}(\theta, M, \alpha)$
\STATE $M \leftarrow M \odot \mathbbm{1}(\theta = 0)$ \quad\COMMENT{freeze non-zero weights}
\STATE $\theta_{\text{dense}}\leftarrow \texttt{densify}(\theta_{\text{sparse}}, M)$
\OUTPUT $\theta_{\text{dense}}$, $M$
\end{algorithmic}
\end{algorithm}

Algorithm~\ref{alg:single} provides pseudo-code for the procedure described above.
Whether any network weight is \emph{frozen} (0) or \emph{learnable} (1)
is indicated by a binary mask, $M$. 
For consistency with later extensions we allow an initial mask as 
optional input to the algorithm.
For single-stage \method, where no weights are frozen initially,
we initialize the mask as a vector of all $1$s (line 1).
The \texttt{sparsify} step (line 2) now calls an arbitrary 
subroutine for network sparsification. 
In the case a non-trivial mask was provided, the sparsification
method must respect the mask, \ie it may not change or remove 
weights indicated as frozen. This is hardly a restriction, though,
as we will discuss below.
After the sparsification, the mask is updated to mark all weights
in the sparse subnetwork, \ie all non-zero ones, as frozen (line 3).
Here, $\odot$ indicates componentwise multiplication, and 
$\mathbbm{1}(P)=1$ if a predicate, $P$, is true and $\mathbbm{1}(P)=0$ 
otherwise, also applied componentwise in our setting.
The other weights remain learnable.

The subsequent \texttt{densify} step can again an arbitrary subroutine
for network training, as long as it respects the mask, \ie does not 
change frozen weights (line 4).
Finally, the algorithm outputs the weights of the dense network, 
$\theta_{\text{dense}}$, together with the updated binary mask, $M$. 

At prediction time, the sparse network can be extracted effortlessly 
from the dense one as $\theta_{\text{sparse}}= (\mathbf{1}-M)\odot\theta_{\text{dense}}$.
Because all its weights were frozen after the sparsification step and 
did not change during the subsequent densification, it is identical 
to the originally embedded one. 

\subsection{Multi-level \method}
Algorithm~\ref{alg:single} takes as input a dense network and a 
binary mask, and it outputs again a dense network and a binary mask.
As such, it is clear that the algorithm can also be run iteratively.
Figure~\ref{fig:multilevel} (c)--(e) illustrates a second 
invocation of \method at a lower sparsity level: the input now 
is a dense network with some frozen (blue) weights (c). 
\texttt{sparsify} produces a sparse network that contains all 
originally frozen weights as well as some additional ones (orange).
These are then frozen, too (d).
\texttt{densify} again produces a dense network from this 
that now contains two embedded sparse networks (e).

\begin{algorithm}[t]
\caption{\texttt{Multi-level \method}}\label{alg:multi}
\begin{algorithmic}[1]
\INPUT network with weights $\theta^{(0)}\in\mathbb{R}^D$
\INPUT target sparsity levels $\alpha^{(1)} > \alpha^{(2)} > \dots > \alpha^{(T)}$
\STATE $M^{(0)} \leftarrow \mathbf{1}\in\mathbb{R}^D, \quad N^{(0)} \leftarrow \mathbf{1}\in\mathbb{R}^D$
\FOR{$t=1,\dots,T$}
  \STATE $\theta^{(t)},M^{(t)} \leftarrow \texttt{\method}(\theta^{(t-1)}, M^{(t-1)}, \alpha^{(t)})$
  \STATE $N^{(t)} \leftarrow N^{(t-1)} + M^{(t)}$
\ENDFOR
\OUTPUT $\theta^{(T)}$, $N^{(T)}$ %$M^{(1)},\dots,M^{(T)}$
\end{algorithmic}
\end{algorithm}

Algorithm~\ref{alg:multi} provides pseudo-code for multi-level \method 
for any number of steps and any decreasing sequence of sparsity levels.
After initializing the mask (line~1), it repeatedly calls \method.
The output of each call serves as input for the next one (line~3). 
Over $T$ rounds, it computes an alternating sequence of \emph{sparse 
and dense models}, 
$\theta^{(1)}_{\text{sparse}}, \theta^{(1)}_{\text{dense}}, \theta^{(2)}_{\text{sparse}}, \dots, \theta^{(T)}_{\text{sparse}}, \theta^{(T)}_{\text{dense}}$,
and well as a sequence of \emph{binary masks}, $M^{(1)},\dots,M^{(T)}$, 
and \emph{aggregate counters}, $N^{(1)},\dots,N^{(T)}$.
Note that the index $(t)$ in Algorithm~\ref{alg:multi} is just
a notational convention. Really, none of the intermediate 
quantities, $\theta$, $M$ or $N$, need to be memorized, but 
they can simply be overwritten with their updated counterpart. 
Consequently, the memory usage is constant, regardless 
of $T$, and the overhead compared to ordinary network training is small.

The use of \method as a subroutine means that each sparse network
could be recovered from the subsequent dense network. For this, one
sets all weights to zero that are not part of the subnetwork, 
\ie one multiplies the dense model's weights with the associate 
binary sparsity mask, 
$\theta^{(t)}_{\text{sparse}}= (\mathbf{1}-M^{(t)})\odot\theta^{(t)}_{\text{dense}}$.
By construction, all weights corresponding to zero entries of
$M^{(t)}$ will be frozen in subsequent steps and therefore remain 
unchanged until the end of the algorithm. 
Consequently, the same sparse network can also be extracted from the 
final dense model as $\theta^{(t)}_{\text{sparse}}= (\mathbf{1}-M^{(t)})\odot\theta^{(T)}_{\text{dense}}$.

Preserving all intermediate masks is not required, either: 
by construction they have a nested structure, such that once a 
mask entry is $0$, it be will also be $0$ in all later masks. 
Consequently, if suffices to store at which time step 
(if any) a weight was frozen for the first time, \ie the
value $t$ for which it first received a $0$ entry in the mask.
This is the role of the counters $N^{(t)}$, which in each step 
are incremented only for those weights that are not frozen (line 4). 
After $T$ steps, binary masks for each sparsity level, $\alpha_t$, 
can be extracted from the final counters $N^{(T)}$ as 
$M^{(t)}=\mathbf{1}-\mathbbm{1}(N^{(T)}\leq t)$.
Weights with counter value $T+1$ are not part of any sparse 
networks but only contribute to the final dense one. 
Ultimately, all sparse networks are available as 
\begin{align}
\theta^{(t)}_{\text{sparse}}= \mathbbm{1}(N^{(T)}\leq t)\odot\theta^{(T)}_{\text{dense}}.
\label{eq:multi}
\end{align}
in identical form to when they were constructed.

Overall, multi-level \method extends single-level \method
to allow one-shot pruning of a single dense network 
to different sparsity levels. At the same time it preserves 
the desirable property that the extracted networks are 
bit-exact identical to the ones stored at training time. 
They are ready-to-use and their performance is known a priori. 

\subsection{Choice of Subroutines}
\method's central components are the subroutines 
\texttt{sparsify} and \texttt{densify}. 
The choice of densification routine is fully flexible, as 
its task is only to improve the accuracy of a given sparse network 
by learning values for the initially zero-valued weights 
without changing the initially non-zero ones. 
Consequently, any standard network training method can be used, 
as long as it allows restricting the set of weights that are 
allowed to change~\cite{ruder2016overview,bottou2018optimization,soydaner2020comparison}. 
In particular, this is the case for standard iterative optimizers, 
such as SGD~\cite{robbins1951stochastic} or Adam~\cite{kingma2014adam}: all one has to do is 
multiply the computed updates by the current mask before adding 
them to the weights. 
In some case, the densification can even be skipped at intermediate
stages. One example of this is if the sparsification routine 
operates via an implicit densification step anyway, as will 
be the case in some of our experiments.

The sparsification routine can also be chosen freely, as long as 
it supports partially sparsifying a network in a way such that 
a given subset of weights must not change.
The latter property holds in particular for most score-based 
methods. 
For example, assume \emph{top-K} magnitude pruning~\cite{han2015learning}. 
For each entry, $\theta_i$, of the network weights 
one computes an \emph{importance score}, $s_i$, for example, 
the weight's absolute value~\cite{hertz1991introduction} 
or its entry in the loss gradient~\cite{karnin1990simple,reed1993pruning}. 
One then keeps the $K$ weights with largest scores and 
sets the others to $0$.
This process has a straight-forward extension to the situation 
in which some weights are frozen: setting $s_i=\infty$ (or 
any large enough finite value) for all frozen weights,
they will fall automatically into the top-$K$ set and thereby 
avoid being removed. 
Note that this scenario applies not only to \emph{unstructured} 
sparsity, but also to \emph{semi-structured} or \emph{structured sparsity}, 
with the scores defined on the right level of granularity~\cite{zhang2022learning,liu2017learning}. 

\subsection{Extension: Overhead-Free \method}
Besides the dense network, Algorithm~\ref{alg:multi} produces the 
counter structure, $N^{(t)}$, which stores which subset of network 
weights should be used at which level of sparsity. 
Storing the counter requires $\tau:=\lceil\log_2 (T+1)\rceil$ bits 
per weight entry.
In practice, however, we can save ourselves even this overhead, 
and dispense with the counter structure completely. 
For this we observe that for compatibility reasons it is common practice 
to store and distribute network weights as 32-bit floating point numbers, 
even if the least significant bits of these values have no effect on the 
network computation, and---at least on GPUs---the actual computations happens 
at a lower precision anyway.\footnote{Trivially so when using low-precision 
data types, such as \texttt{float16} % (10 bit mantissa) 
or \texttt{bfloat16}, %(7 bit mantissa)
but also when using seemingly full-precision \texttt{float32}, because on 
recent GPUs matrix multiplications are executed by default in an 
internal \texttt{TensorFloat32} representations, which uses only a 10-bit 
mantissa rather than the 23-bit of the IEEE standard~\cite{choquette2021nvidia}.}

Consequently, we can store the counter values directly as part of the 
network weights: after each sparsification step, $t$, we overwrite 
the least significant $\tau$ bits of the newly found non-zero weights 
with the binary representation of the value $t$.
We do not change weights that were frozen in previous steps (their least 
significant bits already contain their respective time step), and zero-valued 
weights, \ie weights that are not part of the sparse network, are not 
influenced either. 
In storing the final dense network, we set its least signficant bits 
to $0$, thereby uniquely marking them as not part of any sparse subset.

At deployment time, all that is required to extract a subnetwork of 
desired sparsity level, $\alpha_t$, is to load the dense network and 
keep only those weights with least significant bits $\{1,\dots,t\}$, 
while setting the others to $0$:
\begin{align}
\theta^{(t)}_{\text{sparse}}= 
\mathbbm{1}(\,0<\mathsf{lsb}_{\tau}(\theta^{(T)})\leq t\,)\odot\theta^{(T)}_{\text{dense}},
\label{eq:embedded}
\end{align}
where the function $\mathsf{lsb}_{\tau}$ returns the integer represented 
by the $\tau$ least significant bits of the argument (interpreted as just
a sequence of bits). 
Equation~\eqref{eq:embedded} can easily be implemented using bit masking.

Subsequently, for every practical number of sparsity levels, 
storage requirements for a full \method-net is identical to 
that of a single dense network, and the computational overhead 
for extracting any desired subnetwork is negligible.
The format even allows \emph{streaming} extraction, because 
the decision if a weight should be included or not can be
made immediately after reading its value.
Consequently, the dense model never even has to be stored 
fully on the client device.

\subsection{Extension: Batchnorm Statistics}
Both variants of \method can be applied out-of-the-box for 
networks consisting of arbitrary layers with learnable weights, 
let them be fully-connected, convolutional or attention-based.
However, a different situation emerges for \emph{batch normalization} 
layers. These have not only learnable weights, but also a 
separate set of \emph{batchnorm statistics}, which are determined 
at training time but used only at prediction time. 
Those, typically a \emph{mean} and a \emph{variance} vector,
cannot be pruned and must be adapted to the activation values 
of the preceeding network layer. 
As a consequence, the optimal batchnorm statistics will differ 
between different sparse networks. 

There is a number of possibilities how to add this issue. 
Most straight-forward is to simply store the batchnorm statistics 
for each sparsity level and look them up at deployment time. 
In contrast to the discussion above, this imposes only a minor 
overhead, because the batchnorm statistics are typically several 
orders of magnitude smaller than the set of network weights. 

In a setting of a likely domain shift, suitable batchnorm 
statistics can be computed at prediction time from a subset 
of the prediction time data. 
This is possible because the batchnorm statistics depend only 
on the network activations, not on the loss value, so they can 
be computed also from unlabeled data. 

Finally, in some cases it is possible to remove the batch normalization 
layers all-together from the network by \emph{folding} them 
into the regular network weights. %
This option depends on the chosen architecture though, so we do 
not study it further in this work.

\begin{table}[t]
\caption{Accuracy of \method on CIFAR100 (WideResNet-28-10). 
Models in the caption \emph{ELSA sparse} are extracted from the 
\emph{ELSA dense} model by means of Equation~\ref{eq:embedded}.}\label{tab:CIFAR100-single}
 \centering\addtolength{\tabcolsep}{-2pt}
 \subfloat[global sparsity]{
 \begin{tabular}{c|c|c|c}
  sparsity level  & initial dense & \method sparse & \method dense \\\hline
      80 (\np{79.938385}\%) & $\meannstdd{82.101363}{82.291669}{81.881011}{82.371795}$
                          & $\meannstdd{81.660658}{81.320113}{81.099761}{81.310093}$
                          & $\meannstdd{82.431889}{82.121396}{81.911057}{82.391828}$ \\
      90 (\np{89.930687}\%) & $\meannstdd{82.031250}{82.101363}{82.311696}{82.221556}$
                          & $\meannstdd{80.478764}{80.979568}{80.669069}{81.099761}$
                          & $\meannstdd{81.620592}{82.301682}{82.241589}{82.131410}$ \\
      95 (\np{94.926834}\%) & $\meannstdd{82.151443}{82.071316}{82.131410}{82.892627}$
                          & $\meannstdd{80.038059}{80.018032}{79.977965}{80.208331}$
                          & $\meannstdd{82.141429}{82.081330}{82.271636}{82.441908}$
 \end{tabular}\label{tab:CIFAR100-single-global}}
  \\\medskip
 \subfloat[uniform sparsity]{
 \begin{tabular}{c|c|c|c}
  sparsity level & initial dense & \method sparse & \method dense \\\hline
      80 (\np{79.840042}\%) & $\meannstdd{82.451922}{81.981170}{82.371795}{82.251602}$
                          & $\meannstdd{82.021236}{82.151443}{81.590545}{81.570512}$
                          & $\meannstdd{82.031250}{82.491988}{82.562101}{82.632214}$  \\
      90 (\np{89.930687}\%) & $\meannstdd{82.361782}{82.481968}{81.810898}{82.121396}$
                          & $\meannstdd{81.550479}{81.229967}{81.610578}{81.590545}$
                          & $\meannstdd{82.421875}{82.131410}{82.301682}{82.251602}$  \\
      95 (\np{94.926834}\%) & $\meannstdd{82.071316}{82.091343}{81.870991}{82.321715}$
                          & $\meannstdd{80.889422}{80.679089}{80.729169}{80.518830}$
                          & $\meannstdd{82.491988}{82.522035}{82.371795}{82.421875}$  \\
 \end{tabular}\label{tab:CIFAR100-global-single}}
 \\\medskip
\subfloat[N:M sparsity]{
 \begin{tabular}{c|c|c|c}
  sparsity level & initial dense & \method sparse & \method dense \\\hline
      2:4 (\np{49.899994}\%) & $\meanstd{82.041264}{82.251602}{81.951123}$
                           & $\meanstd{82.041264}{81.720752}{81.450319}$ 
                           & $\meanstd{82.632214}{82.321715}{81.951123}$ \\
      1:4 (\np{74.849998}\%) & $\meanstd{81.790864}{82.141429}{82.051283}$ 
                           & $\meanstd{81.440306}{81.360179}{81.390226}$
                           & $\meanstd{81.840944}{82.061297}{82.552081}$ \\
      1:8 (\np{87.324997}\%) & $\meanstd{82.301682}{82.001203}{82.311696}$ 
                           & $\meanstd{80.378604}{80.568910}{80.809295}$
                           & $\meanstd{81.961137}{82.051283}{82.421875}$\\
 \end{tabular} \label{tab:CIFAR100-NM-single}}
 \end{table}

\section{Experiments}\label{sec:experiments}
We report on experiments in two standard setups 
(ResNet50 on ImageNet, WideResNet-28-10
on CIFAR100) to demonstrates that \method 
works well for relevant model and dataset sizes. 
We also perform stress test experiments on CIFAR10 
that aim at pushing \method to its limits. 

In all cases, the focus of our experiments is to 
show that the quality of \method-Nets is comparable 
to that of sparse networks which were trained individually.
This fact establishes that it is possible to benefit 
from the advantages of having a single dense network 
for deployment while suffering no or almost no loss 
of accuracy. 

\myparagraph{Implementation}
Our implementation of ELSA is written in \emph{jax}~\cite{jax}
based on the \emph{flax}~\cite{flax} and 
\emph{jaxpruner}~\cite{jaxpruner} libraries. 
The source code is provided as supplemental material and 
will be make public after publication.

\begin{table}[t]
\caption{Accuracy of \method on ImageNet (ResNet-50).
Models in the caption \emph{ELSA sparse} are extracted 
from the \emph{ELSA dense} model by means of Equation~\ref{eq:embedded}.
} \label{tab:imagenet-single}
 \centering\addtolength{\tabcolsep}{-2pt}
 \subfloat[global sparsity]{
 \begin{tabular}{c|c|c|c}
  sparsity level & initial dense & \method sparse & \method dense \\\hline
      70 (\np{69.851768}\%)  & $\meannstdd{76.387531}{76.428223}{76.289874}{76.383466}$
                           & $\meannstdd{75.972492}{75.937909}{75.921631}{75.795490}$
                           & $\meannstdd{76.159668}{75.687665}{75.848389}{75.640869}$ \\
      80 (\np{79.830589}\%)  & $\meannstdd{76.387531}{76.428223}{76.383466}{76.383466}$
                           & $\meannstdd{75.842285}{75.903320}{75.882977}{75.720215}$
                           & $\meannstdd{76.306152}{76.202393}{76.086426}{76.161700}$ \\
      90 (\np{89.809410}\%)  & $\meannstdd{76.387531}{76.428223}{76.383466}{76.300049}$
                           & $\meannstdd{75.138348}{75.130206}{74.951172}{75.073242}$
                           & $\meannstdd{76.470947}{76.428223}{76.568604}{76.613361}$ \\
 \end{tabular}\label{tab:imagenet-uniform-single}}
 \\\medskip
 \subfloat[uniform sparsity]{
 \begin{tabular}{c|c|c|c}
  sparsity level & initial dense & \method sparse & \method dense \\\hline
      70 (\np{64.315102}\%)  & $\meannstdd{76.387531}{76.428223}{76.383466}{76.383466}$
                           & $\meannstdd{76.057941}{75.852460}{75.832111}{75.895184}$
                           & $\meannstdd{76.074219}{76.080322}{75.980633}{75.950116}$
                           \\
      80 (\np{73.502968}\%)  & $\meannstdd{76.387531}{76.428223}{76.383466}{76.383466}$
                           & $\meannstdd{75.844318}{75.646973}{75.606281}{75.543213}$
                           & $\meannstdd{75.966388}{76.155597}{76.245117}{76.088458}$
                           \\
      90 (\np{82.690849}\%)  & $\meannstdd{76.387531}{76.428223}{76.383466}{76.383466}$
                           & $\meannstdd{74.409992}{74.222821}{74.365234}{74.239093}$
                           & $\meannstdd{76.147461}{76.043701}{76.198322}{76.204425}$
                           \\
 \end{tabular}\label{tab:imagenet-global-single}}
 \\\medskip
\subfloat[N:M sparsity]{
 \begin{tabular}{c|c|c|c}
  sparsity level & initial dense & \method sparse & \method dense \\\hline
      2:4 (\np{45.939362}\%) & $\meannstdd{76.387531}{76.515704}{76.464844}{76.531982}$
                           & $\meannstdd{76.049805}{75.956219}{75.950116}{76.180011}$
                           & $\meannstdd{75.840253}{75.805664}{76.149493}{75.785321}$
                            \\
      1:4 (\np{68.909134}\%) & $\meannstdd{76.387531}{76.515704}{76.464844}{76.531982}$
                           & $\meannstdd{75.238037}{75.038654}{75.012207}{75.075275}$
                           & $\meannstdd{76.013184}{76.017255}{76.521808}{76.192218}$
                           \\
      1:8 (\np{80.393906}\%) & $\meannstdd{76.387531}{76.515704}{76.464844}{76.531982}$
                           & $\meannstdd{72.682697}{72.526044}{72.483319}{72.532147}$
                           & $\meannstdd{75.919598}{76.088458}{76.257324}{76.151532}$
                           \\
 \end{tabular} \label{tab:imagenet-NM-single}}
 \end{table}

\myparagraph{Sparsity types and levels}
Our experiments cover unstructured (\emph{global sparsity} 
and \emph{uniform (per-layer) sparsity}), as well as 
semi-structured \emph{N:M sparsity}~\cite{choquette2021nvidia}. 

\myparagraph{Training}
We train all initial models with standard 
mini-batch SGD with momentum and Nesterov
acceleration. 
Further details, including hyperparameter choices 
are provided in the supplementary material.

As exemplary \texttt{sparsify} method, we 
use \emph{gradual magnitude pruning (GMP)}~\cite{hagiwara1994simple}, 
which has proven to be a strong allrounder 
method that is often on par with more complex 
techniques~\cite{hoefler2021sparsity}.
To ensure that no frozen weights are removed 
during sparsification we change jaxpruner's 
score computation to yield fixed large 
values for all frozen weights. 
As \texttt{densify} routine, we use the 
same procedure as for initial training, 
just with 100 times lower learning rate. 
We ensure that frozen weights stay at 
their given value by multiply the 
weight updates by the binary mask of 
which weights are learnable.

Internally, GMP works by incrementally 
increasing the sparsity over its run 
following a polynomial schedule~\cite{GMP}. 
As such, it has initial densification implicitly 
built in, allowing us to dispense with the explicit 
densification step for the cases of unstructured 
pruning on ImageNet and CIFAR100.

\myparagraph{Comparison to published works}
As mentioned above, \method is not actually 
a new technique for network sparsification, and we do 
not claim that it improves the sparsity-accuracy trade-off. 
Rather, essentially any existing or future technique 
for network sparsification can be integrated seamlessly 
into \method as its \texttt{sparsify} routine. 
Consequently, the experimental results we report for \method 
should not be seen as attempts to \emph{improving over a baseline}, 
but rather as evidence that \method offers previously 
unseen functionality reliably across many 
relevant and state-of-the-art scenarios.

To show that our experiments indeed reflect this setting, 
we provide reference results from the literature in this 
section.
On the one hand, the accuracy of our initial networks indeed 
matches what has been reported previously in the literature 
for vanilla training (standard data augmentation, standard 
cross-entropy objective): approximately 76\% for a ResNet50 
on ImageNet2012~\cite{pytorch,wightman2021resnet}, above 
80\% for a WideResNet-28-10 on CIFAR100~\cite{zagoruyko2016wide}, 
and approximately 94\% for a SpeeedyResNet on CIFAR-10~\cite{howtotrainyourresnet}.
On the other hand, the quality of the sparse networks we obtain 
is also consistent with the literature: sparsifying ResNet50 on 
ImageNet with GMP to a sparsity level of 90\% results in an
accuracy loss of 1-2\%. For 80\% it is below 1\%, and for 70\% 
the loss is less than 0.5\%, if any~\cite{gupta2022complexity}.
For WideResNet-28-10, similar results hold~\cite{peste2021ac}.

\subsection{Results}
In our experimental results we always report the 
mean and standard deviation of the models' accuracy 
for runs with four random seeds.
The sparsity levels in the results tables indicate 
the chosen sparsity levels for the pruned weights. 
The actual percentage of zero weights is provided in 
bracket behind it. This can be smaller, because some 
layers are excluded from sparsification.

\paragraph{Single-level \method}
We first report on results for the standard \method (Algorithm~\ref{alg:single}),
in which a sparse network is embedded transparently 
within a dense one. 
Here, by construction and as a defining feature of \method, 
the quality of the embedded sparse network is identical 
to the one produced by the chosen sparsification technique, 
because they have identical weights. 
Our evaluation therefore concentrates the quality of the 
dense network that \method produces by densifying the 
networks after freezing sparse subnetwork.

Tables~\ref{tab:CIFAR100-single} and \ref{tab:imagenet-single} 
show the results for the different sparsity types and 
sparsity levels.
One can see that, apart from some fluctuations due to 
randomness in the learning process, the dense models 
produced by \method generally matches the accuracy of 
the initial model within less than half a percent. 
Consequently, the advantage of \method-Nets, that one 
can also extract a sparse network from it at prediction 
time, comes with essentially no drawbacks. 

\myparagraph{Multi-level \method}
We now report on results for the multi-level \method
(Algorithm~\ref{alg:multi}), in which multiple sparse 
networks are embedded within a single dense one. 
This scenario best matches our motivating example of 
deploying a network in resource-constrained situations, 
where one wants to be able to adapt the model sparsity 
to the target device. 
As such, our emphasis here lies on the assessing the 
quality of the sparse networks, in particular the 
ones created at later stages, when some of the network
weights had already previously been frozen. 

Tables~\ref{tab:cifar100-multi} and \ref{tab:imagenet-multi} 
show the results. For each sparsity type three networks 
of different sparsity levels are embedded in a single 
dense one. 
In each case we compare the accuracy of the resulting
sparse networks to their respective references. For
the sparse networks, this is the result of running 
sparsification independently for each level, \ie 
without any weights frozen previously. 
For the dense output network, the reference is the
quality of simply training networks from scratch, 
as represented by the initial networks that serve 
as input for \method.

One can see that in all cases multi-level \method 
achieves sparse networks of quality comparable 
with the references.
This once again shows that \method's advantage of 
having all networks embedded in a single parameter 
set comes without major drawbacks. 
On CIFAR100, the quality of the ultimate dense network 
fully matches that of the initial dense model which was 
trained from scratch.
For ImageNet, the final dense accuracy is in some 
cases about 0.5\%-1\% lower than the initial one.
One explanation for this fact is that after the 
last sparsification step a non-negligible subset 
of weights remains frozen and cannot be changed
during densification, thus resulting in a harder 
optimization problem. 
On the other hand, this was also the case in our 
experiments for single-level \method for with low 
sparsity, but there the difference was smaller.
Instead, the effect might also just be an artifact 
of the fact that we use common default hyperparameters 
for all densification steps, despite the fact that the 
presence of different rather high-accuracy frozen 
subnetworks can be expected to change the loss landscape. 
If the latter is indeed the case, we expect that a more 
thorough hyperparameter search will increase the results 
further. 

\myparagraph{Stress test}
Finally, we report on a \emph{stress test} experiments
that is meant to explore the scalability of multi-level 
\method to a very large number of embedded network. 
For this, we embed 50 sparse networks within a 
single dense one. 
To avoid excessive computational demands we use the 
SpeedyResNet architecture~\cite{howtotrainyourresnet} 
on the CIFAR10 dataset here. 
The sparsity levels range from $99\%$ to $50\%$, 
corresponding to networks with approximately 50.000 
to 2.4 millions non-zero weights out of the original 
4.7 million. 

\begin{figure*}[t]
\subfloat[global sparsity]{
\includegraphics[width=.5\textwidth]{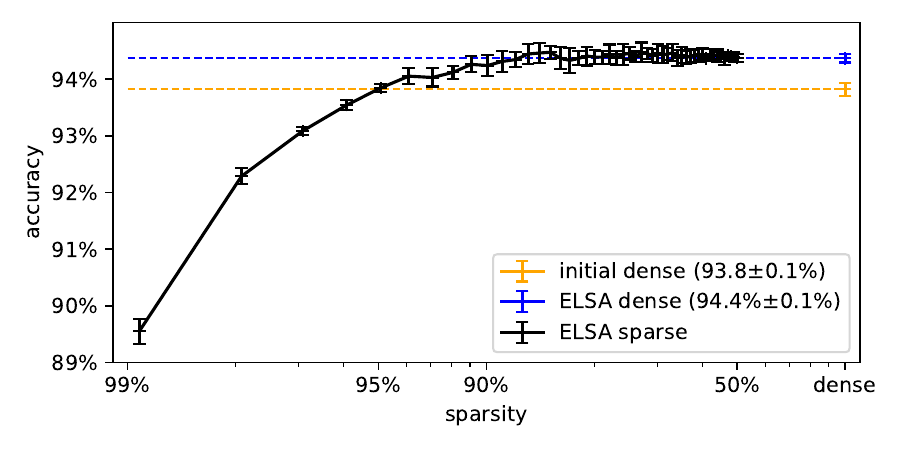}
\label{fig:CIFAR10-global}}
\subfloat[uniform sparsity]{
\includegraphics[width=.5\textwidth]{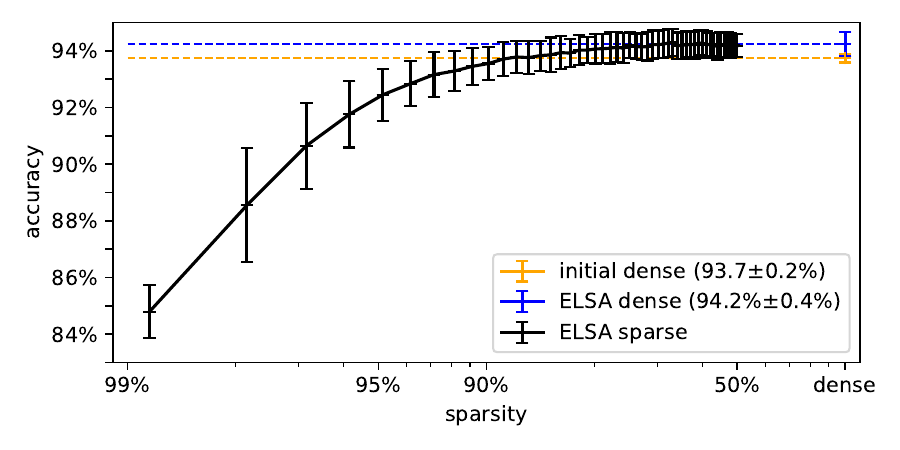}
\label{fig:CIFAR10-uniform}}
\caption{\emph{Stress test experiments.} Multi-level \method is used to create 50 sparse models (CIFAR10, SpeedyResNet) embedded in a single dense one. $x$-axis: model sparsity (logarithmic in number of non-zero model parameters). $y$-axis: accuracy (mean and standard deviation). The dashed horizontal lines illustrate the accuracy of the initial dense network (orange) and the dense output of \method (blue), respectively.}\label{fig:cifar10-multi}
\end{figure*}

Figure~\ref{fig:cifar10-multi} illustrates the results:
one can see that despite the large number of sparsity 
levels \method's performance did not degenerate. 
As it should be, the embedded sparse networks achieve 
(up to random fluctuations) increasing accuracy with 
decreasing sparsity. 
For sparsity levels below a certain threshold 
(approximately 95\% for global sparsity, 
approximately 90\% for uniform sparsity), 
the accuracy of the sparse models matches 
or even exceeds the one of the initial dense 
model, as does the quality of the final dense 
model produced by \method.
Note that this effect allows for an intuitive 
explanation, by observing that the final model 
has undergone many more training epochs than the 
initial one. 
Nevertheless, we find it a noteworthy finding, 
because it demonstrates that it is possible for 
the advantage of deploying variable-sparsity \method-Nets 
not only to come without loss of prediction quality, 
but even with a gain. 

\begin{table}[t]
 \caption{Accuracy of multi-level \method on CIFAR100. All models in the \emph{multi-level \method} column are extracted from the same single dense network by means of Equation~\eqref{eq:embedded}.
Models in the \emph{reference} column are the results of sparsifying independently for each level.}\label{tab:cifar100-multi}
\centering
 \subfloat[global sparsity]{
 \begin{tabular}{c|c|c}
  sparsity level & multi-level \method & reference \\\hline
      95 (\np{94.926834}\%) & $\meannstdd{79.987979}{80.118191}{79.747593}{80.408657}$ % Nov16
                            & $\meannstdd{80.038059}{80.018032}{79.977965}{80.208331}$
      \\
      90 (\np{89.930687}\%) & $\meannstdd{81.009614}{81.169873}{81.330127}{81.340146}$
                            & $\meannstdd{80.478764}{80.979568}{80.669069}{81.099761}$
      \\
      80 (\np{79.938385}\%) & $\meannstdd{81.330127}{81.440306}{81.370193}{81.550479}$
                            & $\meannstdd{81.660658}{81.320113}{81.099761}{81.310093}$
      \\
      dense                 & $\meannstdd{82.241589}{81.810898}{82.161456}{82.021236}$ 
                            & $\meannstdd{82.111377}{81.971157}{82.131410}{82.191509}$
 \end{tabular}}
 \\\medskip
 \subfloat[uniform sparsity]{
 \begin{tabular}{c|c|c}
 sparsity level  & multi-level \method & reference \\\hline
      95 (\np{94.809990}\%) & $\meannstdd{80.468750}{81.069714}{80.869389}{80.228364}$ 
                            & $\meannstdd{80.889422}{80.679089}{80.729169}{80.518830}$
                            \\
      90 (\np{89.819992}\%) & $\meannstdd{81.420273}{81.690705}{81.430286}{81.650639}$
                            & $\meannstdd{81.550479}{81.229967}{81.610578}{81.590545}$
                            \\
      80 (\np{79.839996}\%) & $\meannstdd{81.470352}{81.840944}{81.800884}{82.011217}$
                            & $\meannstdd{82.021236}{82.151443}{81.590545}{81.570512}$
                            \\
      dense                 & $\meannstdd{82.001203}{82.251602}{82.131410}{82.321715}$
                            & $\meannstdd{82.161456}{82.411861}{82.041264}{82.381809}$  \\
\end{tabular}}
\\\medskip
 \subfloat[N:M sparsity]{
 \begin{tabular}{c|c|c}
 sparsity level & multi-level \method & reference \\\hline
      \texttt{1:8} (\np{87.324997}\%) & $\meannstdd{80.548877}{80.799282}{80.528843}{80.578929}$ 
                                      & $\meanstd{80.378604}{80.568910}{80.809295}$ \\ % Nov15 
      \texttt{1:4} (\np{74.849998}\%) & $\meannstdd{81.209934}{81.590545}{81.219953}{81.520432}$ 
                                      & $\meanstd{81.440306}{81.360179}{81.390226}$ \\ % Nov15
      \texttt{2:4} (\np{49.899994}\%) & $\meannstdd{81.951123}{81.820911}{81.520432}{81.640625}$ 
                                      & $\meanstd{82.041264}{81.720752}{81.450319}$ \\  % Nov15
      dense                           & $\meannstdd{82.051283}{82.391828}{82.121396}{82.411861}$
                                      & $\meannstdd{82.081330}{81.830931}{81.991184}{82.491988}$  % Nov15
 \end{tabular}}\\
\end{table}

\section{Related Work}\label{sec:related}
To our knowledge, \method is the first method that 
allows extracting multiple fully trained sparse 
networks at deployment time from a single dense network. 
Consequently, there is no direct prior work in the 
literature with which we could compare quantitatively. 
There are, however, a number of existing methods that 
either perform similar steps for different purposes, 
or that aim for related goals. In this section we 
discuss how \method relates to some of these. 

\myparagraph{One-shot pruning}
\emph{One-shot pruning} refers to the task of creating 
sparse models from dense ones in a single step, \ie, 
without further fine-tuning or other iterative procedures. 

For unstructured sparsity, the state of the  art is 
\emph{CrAM}~\cite{peste2022cram} and its variants.
It proposes a specific loss function, inspired by 
SAM~\cite{foret2020sharpness}, that encourages 
weight values that result in networks of high 
accuracy even when top-K pruned.
By a special optimization procedure it can be optimized
approximately.
A shortcoming of CrAM is that the sparse networks 
it produces are trained only implicitly, so their 
actual quality is not known a priori. 
CrAM also cannot construct batchnorm statistics for them. 
Instead, these have to be created at prediction time, \eg 
from test data. 

\begin{table}[t]
\caption{Accuracy of multi-level \method on ImageNet. All models 
in the \emph{multi-level \method}  column are extracted from the 
same single dense network by means of Equation~\eqref{eq:embedded}. 
Models in the \emph{reference} column are the results of sparsifying 
independently for each level.}\label{tab:imagenet-multi}
 \centering
 \subfloat[global sparsity]{
 \begin{tabular}{c|c|c}
  sparsity level & \method & reference \\\hline
      90 (\np{89.809410}\%) & $\meannstdd{75.205487}{75.278729}{75.276691}{75.268555}$ 
                            & $\meannstdd{75.138348}{75.130206}{74.951172}{75.073242}$
                            \\
      80 (\np{79.830589}\%) & $\meannstdd{75.935870}{76.112872}{75.805664}{75.773114}$
                            & $\meannstdd{75.842285}{75.903320}{75.882977}{75.720215}$ 
                            \\
      70 (\np{69.851768}\%) & $\meannstdd{75.889081}{76.076251}{75.793457}{75.712079}$
                            & $\meannstdd{75.972492}{75.937909}{75.921631}{75.795490}$
                            \\
      dense                 & $\meannstdd{76.214600}{76.094562}{75.911456}{75.604248}$ 
                            & $\meannstdd{76.387531}{76.440430}{76.289874}{76.300049}$
                            \\
 \end{tabular}}
 \\\medskip
 \subfloat[uniform sparsity]{
 \begin{tabular}{c|c|c}
 sparsity level  & \method  & reference \\\hline
      90 (\np{82.690849}\%) & $\meannstdd{74.348956}{74.298096}{74.047852}{74.169922}$
                            & $\meannstdd{74.409992}{74.222821}{74.365234}{74.239093}$
                            \\
      80 (\np{73.502960}\%) & $\meannstdd{75.492352}{75.315350}{75.382489}{75.335693}$
                            & $\meannstdd{75.844318}{75.646973}{75.606281}{75.543213}$
                            \\
      70 (\np{64.315102}\%) & $\meannstdd{75.659180}{75.636798}{75.571698}{75.531006}$ 
                            & $\meannstdd{76.057941}{75.852460}{75.832111}{75.895184}$
                            \\
      dense                 & $\meannstdd{75.724286}{75.742596}{75.793457}{75.856525}$
                            & $\meannstdd{76.387531}{76.428223}{76.383466}{76.383466}$ 
 \end{tabular}}
 \\\medskip
 \subfloat[N:M sparsity]{\label{tab:matryoshka_imagenet_nm}
 \begin{tabular}{c|c|c}
 sparsity level & \method & reference \\\hline
      \texttt{1:8} (\np{80.393906}\%) & $\meannstdd{72.831219}{72.389728}{72.540283}{72.698975}$
                                      & $\meannstdd{72.682697}{72.526044}{72.483319}{72.532147}$
                                      \\
      \texttt{1:4} (\np{68.909035}\%) & $\meannstdd{74.613446}{74.542236}{74.373370}{74.570721}$
                                      & $\meannstdd{75.238037}{75.038654}{75.012207}{75.075275}$
                                      \\
      \texttt{2:4} (\np{45.939358}\%) & $\meannstdd{75.398761}{75.266522}{75.211591}{75.254315}$
                                      & $\meannstdd{76.049805}{75.956219}{75.950116}{76.180011}$
                                      \\
      dense                           &  $\meannstdd{75.944012}{75.773114}{75.551349}{75.724286}$
                                      &  $\meannstdd{76.407880}{76.517743}{76.479083}{76.521808}$
 \end{tabular}}
\end{table}

The empirical results reported by CrAM resemble ours: 
$75.8\%$ and $74.7\%$ accuracy for a ResNet50 trained 
on ImageNet at $80\%$ and $90\%$ sparsity, respectively. 
These numbers cannot be directly compared to \method, 
though, because CrAM uses a stronger (SAM-like~\cite{kwon2021asam}) 
training objective and tunes the batchnorm statistics 
at prediction time.\footnote{In preliminary experiments 
we observed that using the SAM objective for \method leads
to an increase of 1-2\% accuracy in the ImageNet setting
(dense and sparse). We do not include such experiments 
in this work, because they are orthogonal to our main 
contribution.}

Other recent works include \emph{SFW-pruning}~\cite{miao2021learning} 
and \emph{compression-aware SFW}~\cite{zimmer2022compression}. 
Both use Frank-Wolfe optimization to learn networks 
whose weights lie in the convex hull of sparse basis 
vectors, and should therefore be easier approximateable 
in a sparse way. 
Empirically, however, the approaches achieved a worse  
sparsity-accuracy tradeoff than, for example, CrAM, 
and their performance was not demonstrated on 
challenging datasets, such as ImageNet. 

For structured sparsity, OTO~\cite{chen21oto} 
partitions the weights of a network into specific 
groups. It identifies which groups do not contribute 
to the network output and removes those. 
As such, OTO is less flexible than, \eg, score-based
methods, and its effectiveness depends on the 
underlying architecture. In particular, only 
modest sparsity levels were reported for standard 
ResNets.

The OFA~\cite{cai2019once} approach is very broad 
in the type of structures that can be varied, including 
model depth, layer width, kernel size, and input resolution. 
At training time it learns a large model that encompasses 
all of these aspect. At prediction time, a step of 
neural architecture search is performed to find the
best substructure for the current setting. 
A downside of OFA is that it requires the interaction
of several non-standard components, which results in 
a complex training and deployment process. 

\method differs fundamentally from all of the above 
methods: due to it modular structure it is highly
versatile and not restricted to a specific loss 
function (such as CrAM), optimization technique 
(as OFA and the SFW methods) or network architectures 
(such as OTO). 
Also, all the above methods train with via proxy tasks.
The actual sparse networks are created for the first 
time at prediction time. 
\method instead optimizes the sparse networks directly 
at training time. Therefore, it has precise control
over their creation process.
This includes, for example, the computation of suitable 
batchnorm statistics, and the possibility of fine-tuning
the sparse models before freezing and embedding them.

\myparagraph{Alternating sparse and dense training steps}
One characteristic of \method is the alternation of 
sparsification and densification steps. 
Similar patterns have been proposed previously for other purposes.

DSD~\cite{han2016dsd} trains a deep network by alternating 
between sparse phases, in with a sparsity patterns is fixed, 
and dense phases, in which all weights are allowed to change 
again. The goal, however, is not sparsification, but to increase 
the quality of the ultimate dense network.
AC/DC~\cite{peste2021ac} and GaP~\cite{ma2022effective} follow similar patterns, 
but with the goal of producing sparse networks of highest possible accuracy.

The main difference of both methods to \method is that they 
do not preserve the sparse networks they produce during the 
process, but overwrite them. 
Instead of competitors to \method they are rather examples of 
an advanced densification (DSD) and sparsification (AC/DC, GaP) 
techniques that could be used as subroutines within \method.

\myparagraph{Nested network structures}
Having groups of weights within a networks 
that are learned at different times is a 
recurring pattern in the machine learning 
literature.
Because of the space restrictions, we only highlight a few prominent examples here.

The classical \emph{cascade correlation} 
architecture~\cite{fahlman1989cascade} learns
the structure and weights of a (shallow) 
neural network by iteratively adding single
neurons, learning their input weights, and
then freezing those. 
In a continual learning scenario, \emph{progressive networks}~\cite{rusu2016progressive} 
extend previously trained network layers with 
additional neurons in order to continuously 
increase their capacity. Earlier parts of 
the network are frozen to prevent 
catastrophic forgetting.
In the context of network sparsification, 
the \emph{Lottery Ticket Hypothesis}~\cite{frankle2018lottery} 
postulates that dense networks contain sparse subnetworks, 
which can be retrained from scratch to reach similar 
quality levels as the original network. 
FreezeNet~\cite{wimmer2020freezenet} freeze all weights
except a small subset at their random value from initialization.

Most similar to our work is the \emph{PackNet} framework~\cite{mallya2018packnet}.
It follows the same principle of sparsifying a network, freezing 
the non-zero weights and then training the remaining weights again. 
However, it does so in the context of continual multi-task learning, 
using different datasets for the different training phases. 
Network sparsification is used as a tool to free up capacity 
in the network, thereby allowing it to learn new tasks. 

None of these methods share \method's focus on easy storage 
and deployment of explicitly trained sparse models.

\section{Conclusion}\label{sec:conclusion}
We have presented \method, a technique for 
constructing one or more sparse networks as 
embedded subsets of a single dense network. 
\method is completely flexible in the method 
used for sparsification and densification, in 
particular it puts no restriction on the loss 
function, the architecture, or the optimization 
technique.
From the resulting \method-Nets fully trained 
networks of desired sparsity levels can be extracted 
effortlessly at prediction time. Thereby, \method 
vastly reduces the complexity of having to deploy 
different networks depending on the available 
resources, \eg, for mobile devices.

Our experiments showed that \method's iterative
way of producing sparse network results in networks 
of comparable quality to the gold standard of training 
sparse networks separately for each target sparsity level.
The few discrepancies that we did observe provide us 
with promising future research directions.
In cases where the ultimate dense network loses
accuracy compared to its initial counterpart: 
is this an effect of the reduced network flexibility 
when some weights are frozen, or it is simply due to 
suboptimal choices of hyperparameters?
In other cases, where the produced dense network 
was of higher accuracy than the initial one: 
is this purely because it had undergone more training 
epochs?
Or does the alternation of dense and sparse training 
steps together with the integration of frozen subset 
have also more fundamental consequences, for example 
on the loss landscape?
We plan to study these questions in future work.

{
    \small
    \bibliographystyle{ieeenat_fullname}
    \bibliography{ms}
}

\clearpage
\appendix
\section{Appendix -- Experimental Details}\label{app:experiments}
Our experiments use standard benchmarks from the network pruning literature. For details of the 
implementation, see the attached source code.

\begin{table}[b]
\centering
\begin{tabular}{|c|c|c|c|}\hline
    parameter & CIFAR10 (SpeedyResNet) & CIFAR100 (WideResNet-28-10) & ImageNet (ResNet50)\\\hline
    batch size & 512 & 256 & 1024 \\
    learning rate &  $0.2\cdot\frac{\text{batchsize}}{256}$ & $0.1\cdot\frac{\text{batchsize}}{256}$& $0.1\cdot\frac{\text{batchsize}}{256}$ \\
    number of epochs & 30 & 100  & 100\\
    number of warmup epochs & 3 & 3 & 5 \\
    optimizer & SGD & SGD & SGD \\
    momentum & 0.9 & 0.9 & 0.9 \\
    Nesterov acceleration & yes & yes & yes \\
    dropout rate & --- & --- & 0.3 \\
    weight-decay & 0.0005 & 0.0005 & 0.0001 \\
    label smoothing & 0.1 & 0.1 & 0.1 \\
    floating point type & float32 & bfloat16 & bfloat16 \\\hline
\end{tabular}
\caption{Training Hyperparameter}
\end{table}

\subsection{ImageNet}

\myparagraph{Dataset} 
The \emph{ImageNet2012} dataset consists of 1.2 million training examples, which 
we use for training and model selection, and 50.000 validation examples, 
which we use for the final model evaluation. Images have variable sizes.

We use the standard preprocess pipeline of the flax neural network library.
For training, the input images are decoded and a patch is extracted 
using tensorflow's \texttt{tf.image.sample\_distorted\_bounding\_box}
function. These are scaled to $224 \times 224$ resolution and with 50\% 
probability flipped horizontally. 
For evaluation, a center patch is extracted from the decoded images 
and scaled to $224 \times 224$ with no further augmentation steps.
In both cases, the resulting image values are then normalized by subtracting 
the mean and dividing by the standard deviation. 

\myparagraph{Model} 
For ImageNet, use a standard ResNet50 as provided in the flax 
neural network library. It has 25,610,152 parameters, out of which 
53,120 (0.2\%) are the batchnorm statistics. 
This network is known to allow high accuracy, even when trained from 
scratch, at reasonable training and inference cost. 
In line with prior work, we sparsify it to $90\%$, $80\%$ and $70\%$ 
zero weights for unstructured sparsity (global or uniform). 
For semi-structured sparsity we use $1:8 (87.5\%)$, $1:4 (75\%)$ and $2:4 (50\%)$. 

\subsection{CIFAR}

\myparagraph{Datasets} 
The \emph{CIFAR10} and \emph{CIFAR100} datasets each consists of 60.000 training 
examples, which we use for training and model selection, and 10.000 test examples, 
which we use for the finale model evaluation. All images are of size $32\times 32$.

For training, the images are padded by $4$ pixels on each side, and then a 
$32\times 32$ patch is extracted from a random position. These are flipped 
horizontally with 50\% probability. The resulting images are then normalized 
as above. 
For evaluation, no preprocessing is applied except for the normalization. 

\myparagraph{Models}
For CIFAR100, we use a standard WideResNet-28-10 with 
36,554,836 parameters, out of which 17,952 (0.05\%) are 
the batchnorm statistics.
This architecture is known to achieve state-of-the-art
accuracy among ConvNets at modest compute requirements.
It is highly overparametrized, though, such that high sparsification 
ratios are possible. In our experiments, we use $95\%$, $90\%$ and 
$80\%$ for unstructured sparsity. Lower sparsity values tend not 
to lead to higher accuracy for this setting.
For semi-structured sparsity we use the same patterns as above. 

For CIFAR10, we use a \emph{SpeedyResNet} architecture as described
in~\cite{howtotrainyourresnet}. It has 4,658,140 parameters out of 
which 3,584 (0.08\%) are the batchnorm statistics. 
This small architecture is designed particularly for high efficiency 
at training and inference time.
In our implementation one training epoch takes approximately 3 seconds 
on a A100 GPU, and a complete training run of 30 epochs less than 2 minutes.
That allows us to perform experiments in many different settings, 
including 50 different sparsity levels $(99\%, \dots, 50\%)$.

\end{document}